\documentclass[sigconf]{acmart}
\usepackage{multirow}
\usepackage{balance}
\AtBeginDocument{%
  }

\setcopyright{acmlicensed}
\copyrightyear{2026}  
\acmYear{2026}  
\setcopyright{cc}  
\setcctype{by}  
\acmConference[MM '26]{Proceedings of the 34th ACM International Conference on Multimedia}{November 10--14, 2026}{Rio de Janeiro, Brazil}  
\acmBooktitle{Proceedings of the 34th ACM International Conference on Multimedia (MM '26), November 10--14, 2026, Rio de Janeiro, Brazil}  
\acmDOI{10.1145/3767308.3835074}  
\acmISBN{979-8-4007-2213-4/2026/11}  

\begin{document}

\title{PoseAdapter: Dual-Stream 2.5D Controllable Image Generation for Complex Multi-Object Scenes}
\renewcommand{\shorttitle}{PoseAdapter: Dual-Stream 2.5D Controllable Image Generation}


\author{Yufeng Chi}
\email{cyf23@mails.tsinghua.edu.cn}
\affiliation{%
  \institution{Tsinghua University}
  \city{Beijing}
  \country{China}
}

\author{Huimin Ma}
\affiliation{%
  \institution{National University of Defense
Technology}
  \city{Hunan}
  \country{China}
}

\author{Fan Gao}
\affiliation{%
  \institution{Tsinghua University}
  \city{Beijing}
  \country{China}
}

\author{Zhice Niu}
\affiliation{%
  \institution{Tsinghua University}
  \city{Beijing}
  \country{China}
}

\author{Keqin Li}
\affiliation{%
  \institution{Tsinghua University}
  \city{Beijing}
  \country{China}
}

\author{Jianmin Li}
\authornotemark[1]
\email{lijianmin@mail.tsinghua.edu.cn}
\affiliation{%
  \institution{Tsinghua University}
  \city{Beijing}
  \country{China}
}

\renewcommand{\shortauthors}{Yufeng Chi et al.}

\begin{abstract}
  While Text-to-Image (T2I) diffusion models have achieved remarkable success, precise spatial and orientational control in multi-object scenes remains a persistent challenge. Existing methods either rely on computationally expensive dense 3D maps or suffer from severe attribute leakage and ``cut-and-paste'' artifacts. To address these limitations, we propose \textbf{PoseAdapter}, a lightweight framework for high-fidelity 2.5D controllable image generation. Instead of dense spatial maps, it establishes precise spatial-angular anchors using an efficient condition layout: individual object captions, 2D bounding boxes, and 3D angles. To resolve the generative trade-off between strict instance isolation and global coherence, we introduce a Context-Aware Dual-Stream Representation. By injecting local object tokens and relation-enriched scene tokens into the visual stream of modern MM-DiT architectures via parallel masked and unmasked pathways, PoseAdapter eliminates attribute leakage while preserving natural inter-object relationships and scene-level coherence. To support this paradigm, we construct \textit{OrientLayout}, a high-quality dataset featuring standardized 2.5D annotations and instance-level decoupled semantics. Extensive experiments demonstrate that PoseAdapter outperforms state-of-the-art baselines in spatial accuracy, orientational precision, and multi-object visual fidelity. Code and dataset will be available at \url{https://github.com/cyf23/PoseAdapter}.
\end{abstract}


%
\begin{teaserfigure}
  \includegraphics[width=0.9\textwidth]{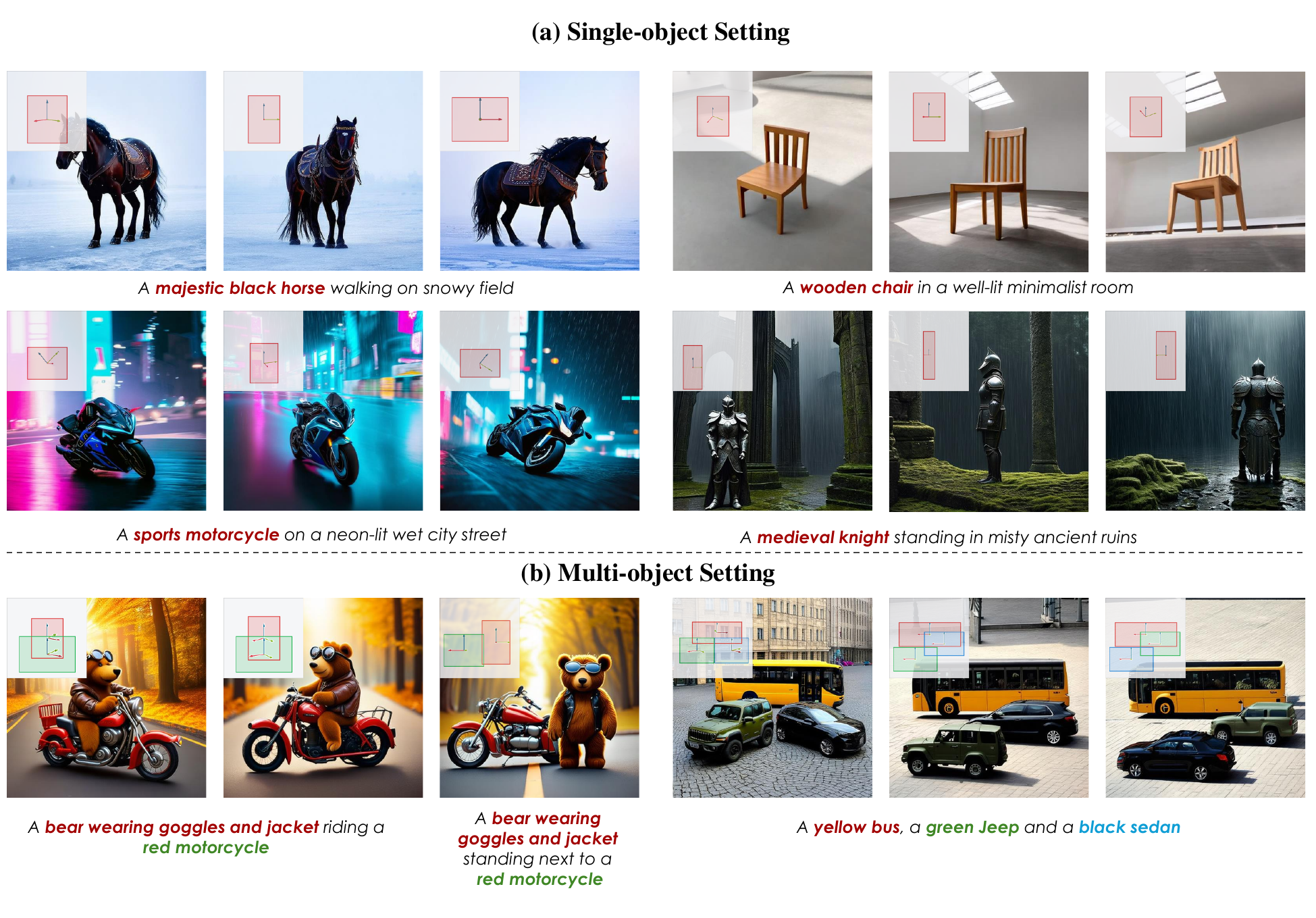}
  \caption{High-fidelity 2.5D controllable generation by \textbf{PoseAdapter}. (a) \textbf{Single-object Setting}: Precise adjustment of 3D orientations (azimuth, polar, rotation) and spatial bounding boxes. (b) \textbf{Multi-object Setting}: Strict attribute decoupling in complex scenes, assigning specific identities and colors to distinct entities while eliminating attribute leakage.}
  \Description{(a) Single-object generation controlled by 3D angles and 2D bounding boxes. (b) Multi-object scenes demonstrating precise attribute assignment without semantic blending.}
  \label{fig:main}
\end{teaserfigure}

\maketitle

\section{Introduction}

Recent advances in Text-to-Image (T2I) diffusion models have greatly expanded the capabilities of visual content creation, enabling the generation of realistic and diverse images from natural language descriptions \cite{rombach2022high, SDXL, SD35}. However, text prompts alone lack the spatial granularity required for complex, real-world multimedia applications. Users often struggle to dictate the exact placement, size, and facing direction of individual entities within a scene. Precise spatial and orientational control has therefore become an important research direction for closing the gap between semantic intent and fine-grained visual output.

\begin{table}[t]
  \caption{Comparison of orientation-aware image generation paradigms. ``Arch. Agnostic'' denotes seamless integration with standard T2I architectures. ``Lightweight Cond.'' indicates that the method does not require dense spatial maps (e.g., depth or CNOCS maps) as input conditions.}
  \label{tab:comparison}
  \centering
  \resizebox{\columnwidth}{!}{
  \begin{tabular}{l c c c c}
    \toprule
    \textbf{Method} & \textbf{Arbitrary 3D Orient.} & \textbf{Lightweight Cond.} & \textbf{Real-World Domain} & \textbf{Arch. Agnostic} \\
    \midrule
    Zero-1-to-3 \cite{zero123} & $\checkmark$ & $\times$ & $\times$ & $\checkmark$ \\
    CompassControl \cite{compasscontrol} & $\times$ & $\checkmark$ & $\times$ & $\checkmark$ \\
    ORIGEN \cite{minorigen} & $\checkmark$ & $\times$ & $\checkmark$ & $\times$ \\
    LOOSECONTROL \cite{loosecontrol} & $\times$ & $\times$ & $\checkmark$ & $\checkmark$ \\
    SceneDesigner \cite{qin2025scenedesigner} & $\checkmark$ & $\times$ & $\checkmark$ & $\checkmark$ \\
    \midrule
    \textbf{PoseAdapter (Ours)} & \textbf{$\checkmark$} & \textbf{$\checkmark$} & \textbf{$\checkmark$} & \textbf{$\checkmark$} \\
    \bottomrule
  \end{tabular}
  }
\end{table}
However, high-fidelity controllable generation remains challenging due to the limitations of existing conditioning methods (Table \ref{tab:comparison}). First, current approaches struggle to balance orientation control, condition efficiency, and real-world generalization. For instance, Zero-1-to-3 \cite{zero123} achieves orientation control via novel view inference but cannot handle complex contextual scenes. CompassControl \cite{compasscontrol} allows lightweight inputs but is restricted to single-angle manipulation and relies on synthetic data, thereby limiting real-world generalization. To achieve comprehensive layout and orientational control in the wild, state-of-the-art frameworks \cite{loosecontrol, qin2025scenedesigner} rely on dense, computationally expensive representations, such as depth or CNOCS maps, which are cumbersome for interactive editing. Furthermore, ORIGEN \cite{minorigen} enables precise 3D grounding without dense maps; however, it relies on optimization with a one-step flow model, preventing integration with standard multi-step diffusion.

Even when spatial and orientational grounding is achieved, orchestrating complex multi-object scenes introduces a second major challenge: preventing attribute leakage. When handling multiple instances, generative models frequently suffer from semantic entanglement, where the color, texture, or identity of one object bleeds into another. To mitigate this issue, a common strategy is to enforce strict spatial masks or bounded attention regions to isolate the generation of each object \cite{compasscontrol,densediffusion,multidiffusion,boxdiff,instancediffusion}. However, this hard-boundary isolation inevitably leads to visible ``cut-and-paste'' artifacts. Without access to neighboring context, the network fails to capture inter-object relationships, thereby losing natural occlusion, relative depth, and consistent lighting. Consequently, resolving the trade-off between strict instance decoupling and global contextual harmony remains an open problem.

To address these challenges, we propose \textbf{PoseAdapter}, a novel framework designed for precise and decoupled multi-object generation. As shown in Figure \ref{fig:main}, our method effectively handles complex spatial placements and dynamic viewpoint shifts in single-object generation, while strictly enforcing attribute decoupling to prevent semantic blending in multi-object compositions.

PoseAdapter achieves this through two core designs. First, we introduce an efficient 2.5D hybrid condition layout for precise spatial and orientational control without dense 3D maps. By parameterizing each instance using a semantic object caption, a 2D bounding box, and three Euler angles, our method defines a spatial-angular anchor for each entity. Second, to resolve the trade-off between strict spatial confinement and natural physical blending, we propose a novel \textbf{Context-Aware Dual-Stream Representation}. Building upon the initial spatial-angular anchors, we encode the multi-modal instance conditions into unified object tokens to establish tight semantic-spatial binding. A scene transformer then models inter-object interactions, producing relation-enriched scene tokens. During the generation phase, these two complementary representations are integrated into the pre-trained diffusion backbone via parallel cross-attention pathways: a masked local stream that enforces strict instance boundaries to eliminate attribute leakage, and an unmasked global stream that preserves inter-object coherence. To adapt to modern MM-DiT architectures \cite{peebles23dit,u-vit,SD35}, we deploy a modality-decoupled injection strategy targeting exclusively the visual stream, keeping textual pathways intact.

To support the training and evaluation of 2.5D controllable generation, we construct \textit{OrientLayout}, a large-scale dataset with fine-grained spatial and orientational annotations. Extensive experiments demonstrate that PoseAdapter outperforms existing state-of-the-art layout and orientation-guided generation methods in visual quality, semantic alignment, and spatial and orientational precision, especially in complex multi-object scenarios.

In summary, our main contributions are as follows:
\begin{itemize}
    \item We propose \textbf{PoseAdapter}, a lightweight and effective framework that achieves precise multi-object T2I generation conditioned on a 2.5D layout (object captions, 2D bounding boxes, and 3D angles).
    \item We introduce a novel \textbf{Context-Aware Dual-Stream Representation} to resolve the multi-object generative trade-off. By utilizing parallel masked and unmasked pathways, this design eliminates attribute leakage while preserving natural inter-object relationships and lighting harmony, and adapts naturally to modern MM-DiT architectures.
    \item We construct \textbf{\textit{OrientLayout}}, a large-scale dataset with fine-grained spatial, orientational, and instance-level caption annotations, and propose a robust Binary Orientation Verification protocol. Extensive evaluations validate our method's superiority in spatial accuracy, orientational precision, and overall visual fidelity.
\end{itemize}

\section{Related Work}

\subsection{Layout-Conditioned Generation}
While T2I models~\cite{rombach2022high,SD35,SDXL,seedream,flux.1kontext} excel globally, precise multi-object spatial control remains challenging. Early methods address this by injecting spatial constraints via gated attention or geometric priors~\cite{gligen,layoutdiffusion,geodiffusion}. To mitigate semantic entanglement in complex scenes, some methods manipulate attention maps~\cite{attend-and-excite} or linguistic guidance~\cite{structureddiffusion}, while common layout-conditioned approaches enforce strict spatial masks~\cite{boxdiff,multidiffusion,instancediffusion,densediffusion}. However, this hard-boundary isolation disrupts inter-object relationships, yielding composition artifacts. Others, like Build-A-Scene~\cite{buildascene}, use multi-stage interactive insertion. In contrast, PoseAdapter achieves one-pass harmonious generation, overcoming isolation artifacts via a Context-Aware Dual-Stream architecture.

\subsection{Orientation-Controllable Generation}

Beyond spatial layouts, dictating the exact 3D orientation of individual entities remains important but difficult. Early efforts like LOOSECONTROL~\cite{loosecontrol} leverage ControlNet~\cite{controlnet} for 3D spatial conditioning but struggle to achieve fine-grained angular precision. Lightweight token-injection frameworks such as C3DW~\cite{c3dw} and CompassControl~\cite{compasscontrol} attempt to encode explicit orientation priors, but their heavy reliance on synthetic training data limits generalization to diverse real-world scenes.
With the emergence of orientation foundation models (e.g., OrientAnything~\cite{orientanything,wangorient}), recent work explores more generalized 3D grounding. Building upon this, ORIGEN~\cite{minorigen} enables zero-shot orientation control via test-time optimization, but suffers from high inference cost due to iterative sampling. SceneDesigner~\cite{qin2025scenedesigner} uses these foundation models to construct large-scale datasets and trains a model conditioned on dense CNOCS maps~\cite{nocs}. While providing explicit geometric guidance, such dense maps are cumbersome for interactive editing and remain prone to geometric deviations in complex compositions.
To balance orientational precision and condition efficiency, PoseAdapter introduces a lightweight 2.5D hybrid layout (object captions, bounding boxes, and three angles), establishing spatial-angular anchors without the overhead of dense 3D maps or iterative optimization.

\begin{figure*}[t]
  \centering
  \includegraphics[width=\textwidth]{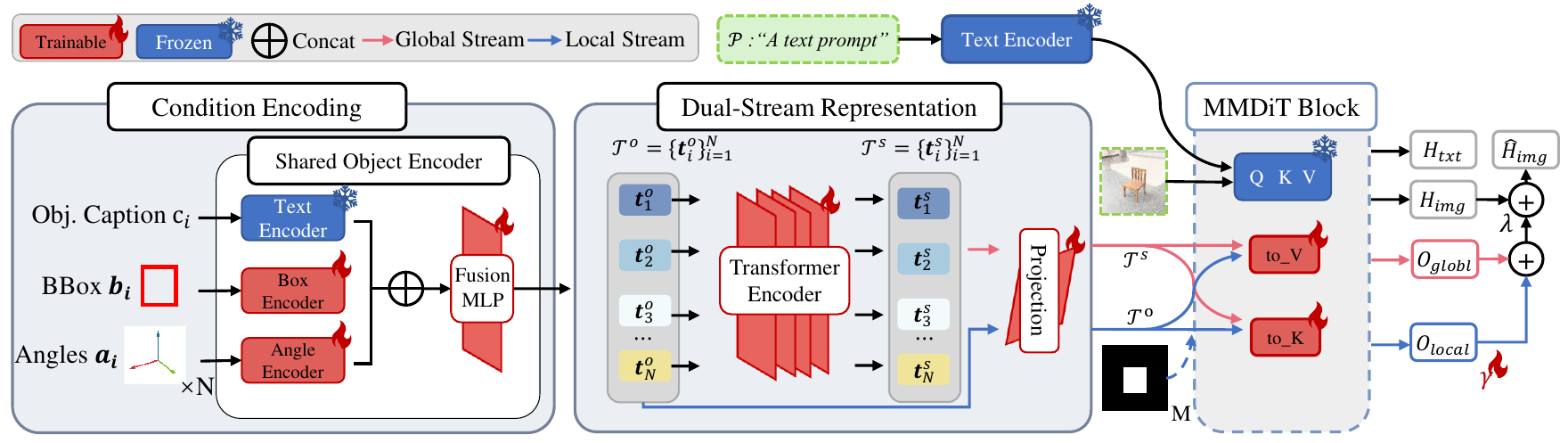}
  \Description{The overall architecture diagram of the proposed PoseAdapter pipeline. On the left, Condition Encoding processes object captions, 2D bounding boxes, and angles through respective encoders and a fusion MLP to generate object tokens. In the middle, a Shared Object Encoder (Transformer) converts these local tokens into relation-enriched scene tokens, establishing a Dual-Stream Representation. On the right, within an MM-DiT block, the visual hidden states interact with these tokens via two parallel cross-attention branches: a Local Stream (using layout masks) for the object tokens, and an unmasked Global Stream for the scene tokens. The outputs are aggregated and residually added to the visual stream.}
  \caption{Overall pipeline of the proposed PoseAdapter. Given a set of instance conditions (captions, 2D bounding boxes, and 3D angles), we first encode them into unified object tokens $\mathcal{T}^o$. A scene transformer then models their interactions to produce relation-enriched scene tokens $\mathcal{T}^s$. Finally, these dual-stream representations are exclusively injected into the pre-trained MM-DiT visual stream via a masked local stream (enforcing precise boundaries) and an unmasked global stream (preserving inter-object coherence).}
  \label{fig:pipeline}
\end{figure*}

\section{Method}
Figure~\ref{fig:pipeline} illustrates the architecture of PoseAdapter. To achieve precise 2.5D multi-object generation, we propose a modality-decoupled, dual-stream injection framework. First, multi-modal instance conditions (comprising semantic captions, 2D bounding boxes (BBoxes), and 3D angles) are encoded into unified object tokens for strict attribute-layout binding (Section~\ref{sec:condition_encoding}). Next, to resolve the multi-object generative trade-off, we introduce a context-aware dual-stream representation that extracts relation-enriched scene tokens while preserving precise local object tokens (Section~\ref{sec:dual_stream_representation}). These two complementary representations are then exclusively injected into the pre-trained MM-DiT visual backbone via parallel cross-attention branches, preventing attribute leakage while maintaining scene-level coherence (Section~\ref{sec:gated_injection}). Finally, we present our layout-aware Flow Matching training objective \cite{flowmatching,SD35,flowstraightandfast} (Section~\ref{sec:training_strategy}) and flexible inference strategies (Section~\ref{sec:inference}).

\subsection{2.5D Layout and Attribute Encoding}
\label{sec:condition_encoding}

Building upon foundational layout-guided paradigms \cite{legonet,atiss,debara,maillard2025laconic,unicontrol}, we adopt a lightweight 2.5D hybrid layout that balances fine-grained control precision with user-friendliness. Specifically, a target scene comprising $N$ objects is defined as $\mathcal{O} = \{o_i\}_{i=1}^N$, where each object is parameterized as a condition tuple $o_i = (c_i, \mathbf{b}_i, \mathbf{a}_i)$.
Here, $c_i$ represents the object caption, $\mathbf{b}_i \in \mathbb{R}^4$ denotes the 2D BBox, and $\mathbf{a}_i \in \mathbb{R}^3$ specifies the 3D orientation. These multi-modal components are jointly encoded to map the scene into a set of initial object tokens $\mathcal{T}^o = \{\mathbf{t}^o_i\}_{i=1}^N$, capturing both the semantic attributes and the spatial-angular configurations of each instance.

\noindent\textbf{Object Caption Embedding.} 
To precisely control the exclusive attributes of each instance $o_i$ (e.g., material, color, and action), its corresponding object caption $c_i$ is processed by a frozen pre-trained CLIP text encoder~\cite{clip,openclip}. We extract the pooled text embedding as the caption feature $\mathbf{f}^{\text{caption}}_i \in \mathbb{R}^{D_t}$, where $D_t$ denotes the output dimension. Encoding detailed captions rather than category labels binds fine-grained attributes to specific entities, mitigating semantic entanglement at the condition level.

\noindent\textbf{Box Encoding.} 
The 2D BBox $\mathbf{b}_i \in \mathbb{R}^4$, parameterized by its min-max coordinates $[x_{\text{min}}, y_{\text{min}}, x_{\text{max}}, y_{\text{max}}]$, explicitly dictates the spatial position and size of $o_i$. To capture subtle positional variations, we expand the normalized coordinates via a Fourier position embedding~\cite{nerf} with 16 frequency bands, then process them through a two-layer MLP with SiLU and LayerNorm, yielding the box embedding $\mathbf{f}^{\text{box}}_i \in \mathbb{R}^{256}$.

\noindent\textbf{Angle Encoding.} 
To explicitly control the 3D orientation of $o_i$, we parameterize it via three angles (azimuth, polar, and rotation), denoted as $\mathbf{a}_i \in \mathbb{R}^3$. While 6D rotation representations~\cite{rotation_representation} suit 3D regression, their higher degrees of freedom complicate generative optimization; we therefore directly encode the angles. To preserve periodicity, we map them into a continuous trigonometric space $\mathbf{e}^{\text{angle}}_i = [\sin(\mathbf{a}_i), \cos(\mathbf{a}_i)] \in \mathbb{R}^6$. The resulting 6D feature is processed by a three-layer MLP with SiLU and LayerNorm, yielding the angle embedding $\mathbf{f}^{\text{angle}}_i \in \mathbb{R}^{256}$.

Finally, these multi-modal embeddings are concatenated and projected via a fusion network ($\text{MLP}_{\text{fusion}}$) to construct the initial unified object token:
\begin{equation}
    \mathbf{t}^o_i = \text{MLP}_{\text{fusion}}([\mathbf{f}^{\text{caption}}_i, \mathbf{f}^{\text{box}}_i, \mathbf{f}^{\text{angle}}_i]).
\end{equation}

\subsection{Context-Aware Dual-Stream Representation}
\label{sec:dual_stream_representation}

Given the initial object tokens $\mathcal{T}^o = \{\mathbf{t}^o_i\}_{i=1}^N$, a straightforward approach is to employ them as independent conditions. To prevent attribute leakage, prior layout-to-image methods~\cite{compasscontrol,densediffusion,multidiffusion,boxdiff,instancediffusion} typically enforce strict spatial constraints. Specifically, for each token $\mathbf{t}^o_i$, a spatial mask is derived from its corresponding bounding box $\mathbf{b}_i$. Formally, the joint layout mask $\mathbf{M} \in \mathbb{R}^{L \times N}$ (where $L$ is the visual sequence length) is defined as:
\begin{equation}
    \mathbf{M}_{p, i} = 
    \begin{cases} 
        0, & \text{if } p \in \mathcal{R}(\mathbf{b}_i) \\ 
        -\infty, & \text{otherwise,} 
    \end{cases}
\end{equation}
where $\mathcal{R}(\mathbf{b}_i)$ denotes the region of $\mathbf{b}_i$. This mask acts as an additive bias in cross-attention to explicitly bound the receptive field:
\begin{equation}
    \text{Attn}(\mathbf{Q}, \mathbf{K}^o, \mathbf{V}^o, \mathbf{M}) = \text{Softmax}\left( \frac{\mathbf{Q} (\mathbf{K}^o)^\top}{\sqrt{d}} + \mathbf{M} \right) \mathbf{V}^o,
\end{equation}
where $\mathbf{K}^o$ and $\mathbf{V}^o$ are projected from $\mathcal{T}^o$. 

While this additive mask effectively drives the attention weights outside the BBox to zero, preventing feature bleeding, it processes each entity in strict isolation. Generating objects without inter-object context yields ``cut-and-paste'' artifacts and fails to capture critical spatial relationships.

Drawing on work in scene understanding \cite{legonet,atiss,debara,maillard2025laconic}, we introduce context-aware scene tokens to overcome this isolation. The isolated object tokens $\mathcal{T}^o$ are fed into a multi-layer Scene Transformer~\cite{transformer}, where self-attention mechanisms allow each object to perceive the semantic and geometric configurations of all other entities. This transforms the localized representations into a set of relation-enriched scene tokens $\mathcal{T}^s = \{\mathbf{t}^s_i\}_{i=1}^N$. These scene tokens encode the global layout context, providing a full receptive field that facilitates natural physical blending.

The unmasked global attention of $\mathcal{T}^s$ promotes scene-level coherence. However, relaxed regional constraints reintroduce the risk of attribute leakage. We therefore combine both representations into a Dual-Stream architecture: $\mathcal{T}^s$ captures global context, while $\mathcal{T}^o$ enforces strict local boundaries. This combination establishes a complete condition space for the subsequent injection strategy.

\subsection{Modality-Decoupled Dual-Stream Injection}
\label{sec:gated_injection}

Following IP-Adapter~\cite{ipadapter}, we integrate these context-aware representations exclusively into the visual stream. Modern MM-DiT architectures decouple visual ($\mathbf{H}_{\text{img}}$) and textual ($\mathbf{H}_{\text{txt}}$) hidden states; injecting our dual-stream conditions solely into $\mathbf{H}_{\text{img}}$ avoids interfering with the pre-trained text representations.

Within each transformer block, the normalized visual query $\mathbf{Q}_{\text{img}}$ interacts with our representations via two parallel cross-attention branches. For parameter efficiency and feature alignment, both local object tokens $\mathcal{T}^o$ and global scene tokens $\mathcal{T}^s$ share the same key and value projection matrices ($\mathbf{W}_k^{ip}$ and $\mathbf{W}_v^{ip}$). Notably, to stabilize the cross-attention computation, the projected keys are further normalized via RMSNorm before being explicitly decoupled.

To enforce strict instance boundaries, the masked local stream computes cross-attention using the dynamic layout mask $\mathbf{M}$:
\begin{equation}
    \mathbf{O}_{\text{local}} = \text{Attn}(\mathbf{Q}_{\text{img}}, \text{RMSNorm}(\mathcal{T}^o\mathbf{W}_k^{ip}), \mathcal{T}^o\mathbf{W}_v^{ip}, \mathbf{M}).
\end{equation}

Simultaneously, the unmasked global stream attends to $\mathcal{T}^s$ without spatial restrictions to capture inter-object relationships:
\begin{equation}
    \mathbf{O}_{\text{global}} = \text{Attn}(\mathbf{Q}_{\text{img}}, \text{RMSNorm}(\mathcal{T}^s\mathbf{W}_k^{ip}), \mathcal{T}^s\mathbf{W}_v^{ip}, \mathbf{0}).
\end{equation}

Finally, the complementary outputs are aggregated via a trainable, zero-initialized gating parameter $\gamma$ and residually injected back into the visual hidden states:
\begin{equation}
    \mathbf{\hat{H}}_{\text{img}} = \mathbf{H}_{\text{img}} + \lambda \cdot (\mathbf{O}_{\text{global}} + \gamma \cdot \mathbf{O}_{\text{local}}).
\end{equation}
where $\lambda$ is a continuous scale factor. The updated representation $\mathbf{\hat{H}}_{\text{img}}$ balances precise instance control with global coherence.

\subsection{Training Strategy}
\label{sec:training_strategy}

Our model is trained end-to-end using the Rectified Flow Matching objective~\cite{flowmatching,flowstraightandfast}. The forward probability path linearly interpolates between the clean data latent $\mathbf{x}_0$ and standard Gaussian noise $\epsilon \sim \mathcal{N}(\mathbf{0}, \mathbf{I})$. The intermediate noisy latent at timestep $t \in [0, 1]$ is defined as $\mathbf{x}_t = (1 - t) \mathbf{x}_0 + t \epsilon$. The network is optimized to predict the flow velocity $\mathbf{v} = \epsilon - \mathbf{x}_0$. 

Following~\cite{SD35,logistic}, we sample the timestep $t$ using Logit-Normal Sampling. Standard T2I synthesis centers this distribution ($m=0$) to emphasize intermediate steps, but layout-conditioned generation must prioritize high-noise phases where macroscopic spatial configurations are established; we therefore apply a positive location shift.
Specifically, we sample $u \sim \mathcal{N}(m, s^2)$ with $m = 1.0$ and $s = 1.0$, mapping to the timestep via $t = \sigma(u)$. This shifts the density toward higher noise levels ($t \to 1$), forcing the model to prioritize global spatial grounding over high-frequency details.

To focus the loss more on foreground objects, we introduce a box mask loss. Rather than computing a uniform MSE across the entire spatial resolution, we dynamically up-weight the penalty within the designated BBox regions. Formally, the spatially weighted objective is defined as:
\begin{equation}
    \mathcal{L}_{\text{flow}} = \mathbb{E}_{\mathbf{x}_0, \epsilon, t, \mathcal{P}, \mathcal{C}} \left[ \frac{1}{L} \sum_{j=1}^L \mathbf{W}_j \cdot \left\| \mathbf{v}_\theta(\mathbf{x}_t, t, \mathcal{P}, \mathcal{C}) - \mathbf{v} \right\|_2^2 \right],
\end{equation}
where $L$ is the spatial sequence length, $\mathcal{P}$ represents the full prompt, $\mathcal{C}$ denotes our dual-stream layout conditions, and $\mathbf{v}_\theta$ is the network prediction. The spatial weight vector $\mathbf{W} \in \mathbb{R}^L$ is computed dynamically:
\begin{equation}
    \mathbf{W}_j = 1.0 + (\alpha - 1.0) \cdot \mathbf{M}_{j}^{\text{loss}}.
\end{equation}
Here, $\mathbf{M}^{\text{loss}}$ is a binary mask indicating the union of all BBoxes, and $\alpha > 1$ is a hyperparameter scaling the penalty inside the objects.

Finally, to enable flexible compositional control and support Classifier-Free Guidance (CFG)~\cite{classifier-free} during inference, we implement a decoupled modality dropout strategy. During training, we randomly drop condition modalities—such as the global caption, object captions, or all conditions—to establish an unconditional baseline. This decoupling ensures the model learns to independently attend to global semantics, local attributes, and layout geometries without entanglement.

\subsection{Inference and Controllability}
\label{sec:inference}

During inference, PoseAdapter accepts a global scene prompt $\mathcal{P}$ and instance conditions $\mathcal{O} = \{(c_i, \mathbf{b}_i, \mathbf{a}_i)\}$. To prevent out-of-distribution textual degradation, the full prompt $\mathcal{P}$ is preserved and processed by the pre-trained text encoders. Simultaneously, the local object captions $c_i$ and their geometric configurations ($\mathbf{b}_i$, $\mathbf{a}_i$) are exclusively injected into the visual stream.

In practice, users simply provide object captions and a background description, which are concatenated to form $\mathcal{P}$ without complex layout-related vocabulary. Alternatively, an off-the-shelf LLM can parse a user instruction into $\mathcal{P}$ and $\mathcal{O}$. Users can also adjust CFG scales to balance textual fidelity with layout adherence.

\section{Dataset Construction: OrientLayout}
\label{sec:dataset}

To support the training and evaluation of 2.5D controllable generation, we construct \textbf{OrientLayout}. In total, the dataset comprises 110K high-quality samples, consisting of 62K images derived from COCO~\cite{coco}, 6K from Cityscapes~\cite{cityscapes,cityscapes3d}, and the remaining 42K from Objectron~\cite{objectron}. Each sample is structured as $(\mathbf{I}, \mathcal{P}, \{(c_i, \mathbf{b}_i, \mathbf{a}_i)\})$, comprising the image $\mathbf{I}$, the full prompt $\mathcal{P}$, and a set of object conditions. Each condition contains a local caption $c_i$, a 2D BBox $\mathbf{b}_i$, and 3D angles $\mathbf{a}_i$ (azimuth, polar, and rotation).

\noindent\textbf{Spatial and Orientational Annotation.} We utilize two complementary data pipelines. For the Objectron and Cityscapes subsets, we derive 3D angles from original 3D poses and extract 2D BBoxes via Grounding DINO~\cite{groundingdino}. For the COCO subset, we leverage the provided 2D BBoxes and annotate the missing 3D angles via OrientAnything~\cite{orientanything}. We filter out low-confidence predictions, excessively small instances, and isotropic objects (e.g., spheres).

\noindent\textbf{Standardization and Manual Verification.} We first resolve coordinate conflicts within Objectron, where the definition of a ``canonical front'' varies across categories (e.g., ``cup'' vs. ``book''). We apply strict coordinate transformations to align these native poses with the unified prediction space of OrientAnything. For the COCO subset, human annotators manually verify the annotations, filtering out erroneous orientations and heavily occluded instances to ensure high-quality ground truths.

\noindent\textbf{Relationship-Aware Semantic Annotation.} Finally, we employ an MLLM~\cite{qwen2.5} to generate the full prompt $\mathcal{P}$ and individual object captions $c_i$. To mitigate semantic ambiguity, we explicitly prompt the MLLM to include target categories and exclude distractors for each BBox. This semantic decoupling ensures $c_i$ serves as a precise anchor without attribute entanglement.

\section{Experiments}

\noindent\textbf{Implementation Details.}
We adopt Stable Diffusion 3.5~\cite{SD35} as our pre-trained T2I backbone. PoseAdapter is trained on the constructed OrientLayout dataset for 60 epochs using 7 NVIDIA RTX 5090 (32GB) GPUs. The image resolution is set to $512 \times 512$ with a total batch size of 24. We employ the AdamW optimizer, initializing learning rates at 3e-5 for the condition encoder and scene transformer, 1.5e-5 for the dual-stream adapter, and 1e-3 for the gating parameters. Furthermore, we set the box mask loss weight $\alpha=3$ and the injection scale $\lambda=1$. Further details and additional analyses are provided in the Supplementary Material.

\subsection{Experimental Setup}
\noindent\textbf{Baselines.} 
We compare PoseAdapter against state-of-the-art 3D-aware controllable generation frameworks, selecting LOOSECONTROL~\cite{loosecontrol} and SceneDesigner~\cite{qin2025scenedesigner} as our primary baselines. Other methods discussed in our literature review are excluded from this evaluation: Zero-1-to-3~\cite{zero123} and CompassControl~\cite{compasscontrol} are constrained to single-object manipulation or specific synthetic domains, while ORIGEN~\cite{minorigen} relies on a distinct optimization-based paradigm.

Regarding the condition formats, while PoseAdapter operates directly on lightweight 2.5D parameters, we provide LOOSECONTROL and SceneDesigner with their required dense representations (e.g., depth or CNOCS maps). Although the SceneDesigner manuscript proposes a Disentangled Object Sampling algorithm, its official open-source implementation relies exclusively on a unified global control pathway. To evaluate its multi-object capabilities, we explicitly provide the spatial layout information and inter-object relationships within the full prompt to guide its generation.

\begin{table*}[t]
 \caption{Quantitative comparison on OrientLayout. $Acc_{22.5^\circ}$: automated accuracy (OrientAnything); $Acc_{BOV}$: human verification.}
  \label{tab:main_results}
  \centering
  \resizebox{\textwidth}{!}{
  \begin{tabular}{l ccc cccc cccc cccc}
    \toprule
    \multirow{2}{*}{Method} & \multicolumn{3}{c}{Overall Quality} & \multicolumn{4}{c}{OrientLayout-Single-Front} & \multicolumn{4}{c}{OrientLayout-Single-Back} & \multicolumn{4}{c}{OrientLayout-Multi} \\
    \cmidrule(lr){2-4} \cmidrule(lr){5-8} \cmidrule(lr){9-12} \cmidrule(lr){13-16}
    & FID $\downarrow$ & CLIP $\uparrow$ & O-CLIP $\uparrow$ & $Acc_{ls}$ $\uparrow$ & mIoU $\uparrow$ & $Acc_{22.5^\circ}$ $\uparrow$ & $Acc_{BOV}$ $\uparrow$ & $Acc_{ls}$ $\uparrow$ & mIoU $\uparrow$ & $Acc_{22.5^\circ}$ $\uparrow$ & $Acc_{BOV}$ $\uparrow$ & $Acc_{ls}$ $\uparrow$ & mIoU $\uparrow$ & $Acc_{22.5^\circ}$ $\uparrow$ & $Acc_{BOV}$ $\uparrow$ \\
    \midrule
    LOOSECONTROL & 36.78 & 0.287 & 0.232 & 34.72 & 32.94 & 10.18 & 12.28 & 35.23 & 34.87 & 6.47 & 8.82 & 24.20 & 29.45 & 2.43 & 6.40 \\
    SceneDesigner & 25.10 & \textbf{0.347} & 0.288 & 75.79 & 69.79 & 73.68 & 86.67 & 74.71 & 66.45 & 62.35 & 80.58 & 47.48 & 46.38 & 56.00 & 80.00 \\
    \textbf{PoseAdapter (Ours)} & \textbf{24.60} & 0.344 & \textbf{0.310} & \textbf{97.84} & \textbf{87.96} & \textbf{87.02}& \textbf{95.44} & \textbf{99.40} & \textbf{88.89} & \textbf{71.76} & \textbf{88.82} & \textbf{91.67} & \textbf{81.72} &\textbf{72.70} & \textbf{93.22} \\
    \bottomrule
  \end{tabular}
  }
\end{table*}

\noindent\textbf{Evaluation Datasets.} As the official validation set of SceneDesigner is unreleased, we reconstruct an equivalent benchmark from the COCO validation set following its protocol. Utilizing the pipeline from Section~\ref{sec:dataset}, we manually scrutinize the data to ensure precise ground-truth orientations and employ random sampling to align the size of our dataset with that of the baseline. The final benchmark comprises \textit{OrientLayout-Single} (285 front, 170 back) and \textit{OrientLayout-Multi} (250 multi-object cases).

\noindent\textbf{Evaluation Metrics.} To evaluate visual fidelity, semantic alignment, and spatial precision, we follow the evaluation protocol of SceneDesigner. Specifically, to measure the visual fidelity of the generated images, we report the Fréchet Inception Distance (FID). For the evaluation of global semantic alignment, we compute the standard CLIP score to assess the consistency between the full prompt and the image. Furthermore, to evaluate the fine-grained semantic alignment of individual entities, we additionally compute an object-level CLIP score (O-CLIP). This metric is calculated by cropping the region of each detected object, computing the CLIP similarity between the cropped region and the corresponding object caption, and subsequently averaging the scores across all instances. Finally, to quantify spatial precision, we calculate the mean Intersection over Union (mIoU) and the spatial accuracy ($Acc_{ls}$ at $\text{IoU} > 0.6$) by utilizing Grounding DINO for the detection of objects.

To evaluate orientation accuracy, following the protocol of recent baselines~\cite{qin2025scenedesigner,minorigen}, we employ OrientAnything to estimate the azimuth angles of the generated objects and report the accuracy within a $22.5^\circ$ tolerance ($Acc_{22.5^\circ}$). However, we observe that applying this automated predictor to generated images occasionally yields unstable predictions, despite our conditioning angles being derived from its own pseudo-labels. Relying solely on automated metrics may not fully capture the human-perceived orientation fidelity.

To address this, we introduce \textbf{Binary Orientation Verification}. The OrientAnything benchmark categorizes azimuths into eight discrete directions (e.g., front, front-left). However, asking human evaluators to classify objects into these eight bins introduces severe subjective bias at angular boundaries. To mitigate this, we quantize the continuous ground-truth azimuth into its corresponding target label among the eight directions. We then ask evaluators a simple binary question: \textit{``Is the object facing [Target Direction]?''} The positive response rate defines our Binary Orientation Verification Accuracy ($Acc_{BOV}$). This design reduces human cognitive load while reliably assessing orientation. Details are in the Supplementary Material.

\begin{figure*}[t]
  \centering
  \includegraphics[width=\textwidth]{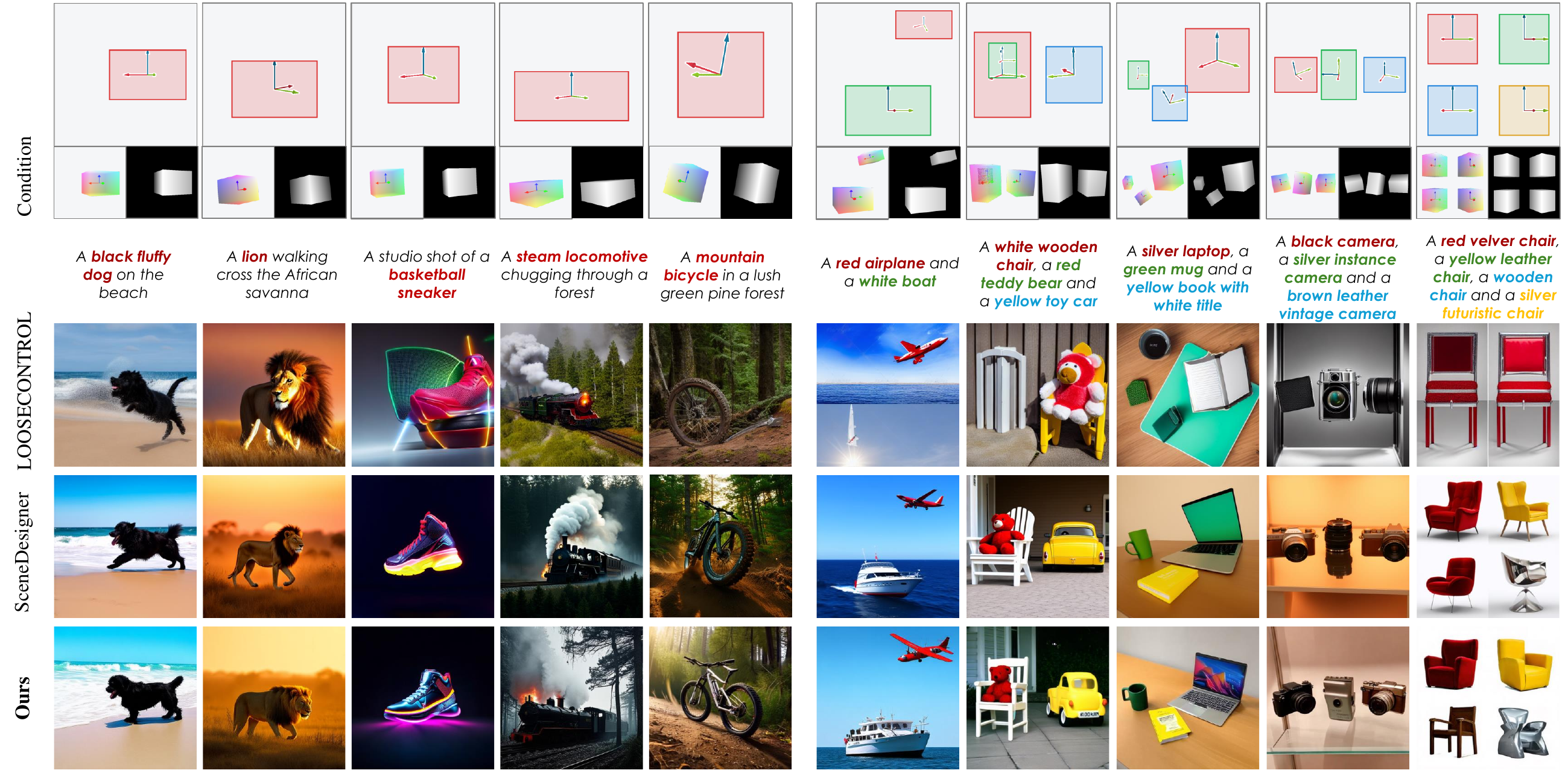} 
  \Description{A large grid of images comparing 2.5D controllable image generation methods. The top rows display the input spatial conditions, showing our lightweight 2D bounding boxes with orientation arrows versus the dense CNOCS and depth maps used by baselines. The grid is divided into columns representing different full prompts, such as a dog, a lion, a locomotive, an airplane, and multi-object scenes like a room with cameras or chairs. The rows show the generated results from LOOSECONTROL, SceneDesigner, and our PoseAdapter. Visually, our results align strictly with the colored bounding boxes and full prompts, while baselines show obvious misalignments, missing objects, or color blending in multi-object columns.}
  \caption{Qualitative comparison of controllable image generation. The proposed PoseAdapter demonstrates precise layout and orientation control across diverse settings. The \textit{Condition} column displays the respective inputs: 2.5D visualizations for the proposed approach, CNOCS maps for SceneDesigner, and depth maps for LOOSECONTROL.}
  \label{fig:qualitative}
\end{figure*}

\subsection{Qualitative Results}

Figure~\ref{fig:qualitative} provides a visual comparison between PoseAdapter and the baseline models across diverse generation scenarios. Specifically, the condition column illustrates the respective spatial and orientational control signals utilized by each method. We display the 2.5D condition visualizations for the proposed approach, the rendered CNOCS maps for SceneDesigner, and the corresponding depth maps for LOOSECONTROL.

\noindent\textbf{Single-object Setting.} LOOSECONTROL merely approximates spatial locations, failing to capture target orientations. SceneDesigner achieves better alignment but remains geometrically imprecise. Specifically, it occasionally violates bounding box constraints (e.g., the ``steam locomotive'') or exhibits azimuth deviations (e.g., the ``basketball sneaker''). In contrast, PoseAdapter consistently synthesizes high-fidelity images complying with the designated layout and 3D orientation.

\noindent\textbf{Multi-Object Setting.} In complex multi-object scenarios, LOOSECONTROL fails to synthesize semantically aligned images. Even after incorporating layout hints into the full prompt and using multiple sampling steps, SceneDesigner still suffers from severe attribute leakage and concept confusion. It visibly tangles the distinct colors and materials across the generated ``cameras'' and ``chairs'', and erroneously dyes the screen of the ``laptop'' green (third column from the right). It also struggles with complex spatial manipulations involving polar and rotation angles (e.g., the ``airplane'' and the central ``camera''). By explicitly binding object captions to 2.5D spatial conditions via our dual-stream architecture, PoseAdapter decouples multiple entities and eliminates attribute leakage, yielding high-fidelity results faithful to both textual semantics and spatial-angular constraints.
\subsection{Quantitative Results}

Table \ref{tab:main_results} summarizes quantitative evaluation results across three dimensions: visual quality, semantic alignment, and spatial/orientational control precision.

\noindent\textbf{Visual Quality and Text-Image Alignment.} PoseAdapter achieves superior photorealism and diversity, as evidenced by the lowest FID. Beyond competitive global semantic alignment, our method demonstrates a clear advantage in object-level CLIP (O-CLIP), consolidating the fact that our dual-stream architecture preserves individual semantic identities and prevents attribute leakage in complex scenes.

\noindent\textbf{Spatial Layout and Orientation Precision.} PoseAdapter consistently outperforms baselines across all splits in fine-grained 2.5D control. LOOSECONTROL struggles with bounding box adherence ($Acc_{ls}$, mIoU) and largely fails in orientation alignment under both automated and human evaluations ($Acc_{22.5^\circ}$, $Acc_{BOV}$), as its coarse dense conditioning lacks accurate boundary enforcement and explicit angular inputs. While SceneDesigner captures general orientations, its spatial precision remains inferior even in single-object scenarios, revealing its incapability to enforce strict scale limits. In the OrientLayout-Multi setting, SceneDesigner's performance drops sharply due to severe inter-object interference.
PoseAdapter leverages explicit 2.5D conditions to achieve the best spatial and orientation accuracy. Notably, while our method leads in automated accuracy ($Acc_{22.5^\circ}$), all methods score systematically lower than in human verification ($Acc_{BOV}$). This gap confirms the instability of automated predictors on generated images, and consolidates the need for $Acc_{BOV}$. Despite performance drops on back-facing objects due to limited back-view data, our method continues to outperform all baselines.

\begin{table}[t]
  \caption{Efficiency and latency comparison (RTX 5090, 3 objects, 28 steps). ``Tr. Params'' denotes trainable parameters; ``Prep.'' indicates external preprocessing time. Our method avoids spatial map rendering entirely.}
  \label{tab:efficiency}
  \centering
  \resizebox{\columnwidth}{!}{
  \begin{tabular}{l l r l c c}
    \toprule
    \textbf{Method} & \textbf{Backbone} & \textbf{Tr. Params} & \textbf{Injection} & \textbf{Prep. (s) $\downarrow$} & \textbf{Net. (s) $\downarrow$} \\
    \midrule
    LOOSECONTROL & SD 1.5 & 0.7M & CN + LoRA & $\sim$14.2 & $\sim$2.6 \\
    SceneDesigner & SD 3.5 & 1,487.7M & ControlNet & $\sim$13.7 & $\sim$2.4 \\
    \textbf{PoseAdapter (Ours)}& SD 3.5 & 228.0M & IP-Adapter & 0.0 & $\sim$3.0 \\
    \bottomrule
  \end{tabular}
  }
\end{table}

\noindent\textbf{Efficiency.}
Using the same SD~3.5 backbone, PoseAdapter requires only 15.3\% of the trainable parameters of SceneDesigner. While our dual-stream cross-attention introduces a marginal overhead during neural inference, our 2.5D representation enables zero-delay condition encoding. Since baselines natively require manual 3D GUI orchestration, we employ an automated 2D-to-3D optimization script to establish a measurable lower bound for their preprocessing. Even so, their intrinsic reliance on 3D engine rendering (e.g., Blender) inflicts an $O(N)$ multi-second bottleneck. By bypassing this entirely, our method yields significantly faster end-to-end generation. Table~\ref{tab:efficiency} provides a detailed quantitative comparison.

\subsection{User Study} 
We conducted a blind user study to acquire human-perceived quality evaluation. To prevent cherry-picking, an LLM was utilized to generate 100 diverse layout instructions (50 single- and 50 multi-object) for categories with distinct canonical fronts. Twenty human evaluators assessed the generated images based on text, spatial, and orientation alignment (Pass/Fail). As reported in Table~\ref{tab:user_study}, PoseAdapter consistently achieves over 90\% approval across all criteria. It significantly outperforms baselines, which suffer severe attribute leakage and geometric distortion, particularly in complex multi-object scenes. Further visualizations and details are deferred to the Supplementary Material.

\begin{table}[t]
  \caption{User study results (approval rate \%).}
  \label{tab:user_study}
  \centering
  \resizebox{\columnwidth}{!}{
  \begin{tabular}{l ccc ccc}
    \toprule
    \multirow{2}{*}{\textbf{Method}} & \multicolumn{3}{c}{\textbf{Single-Object}} & \multicolumn{3}{c}{\textbf{Multi-Object}} \\
    \cmidrule(lr){2-4} \cmidrule(lr){5-7}
    & Text & Orient. & Spatial & Text & Orient. & Spatial \\
    \midrule
    LOOSECONTROL & 45.2 & 15.5 & 38.0 & 16.4 & 3.5 & 7.5 \\
    SceneDesigner & 92.5 & 82.0 & 77.5 & 60.2 & 45.5 & 32.0 \\
    \textbf{PoseAdapter (Ours)} & \textbf{98.5} & \textbf{97.0} & \textbf{98.5} & \textbf{95.0} & \textbf{87.5} & \textbf{95.5} \\
    \bottomrule
  \end{tabular}
  }
\end{table}

\subsection{Ablation Study}
\label{sec:ablation}

To evaluate the core components of PoseAdapter, we ablate our model architecture and training objectives (Table \ref{tab:ablation}). For the architectural ablation, we focus on the OrientLayout-Multi benchmark to perform quantitative comparisons, as single-object scenes insufficiently expose the limitations of isolated local streams.

\begin{figure}[t]
  \centering
  \includegraphics[width=0.98\linewidth]{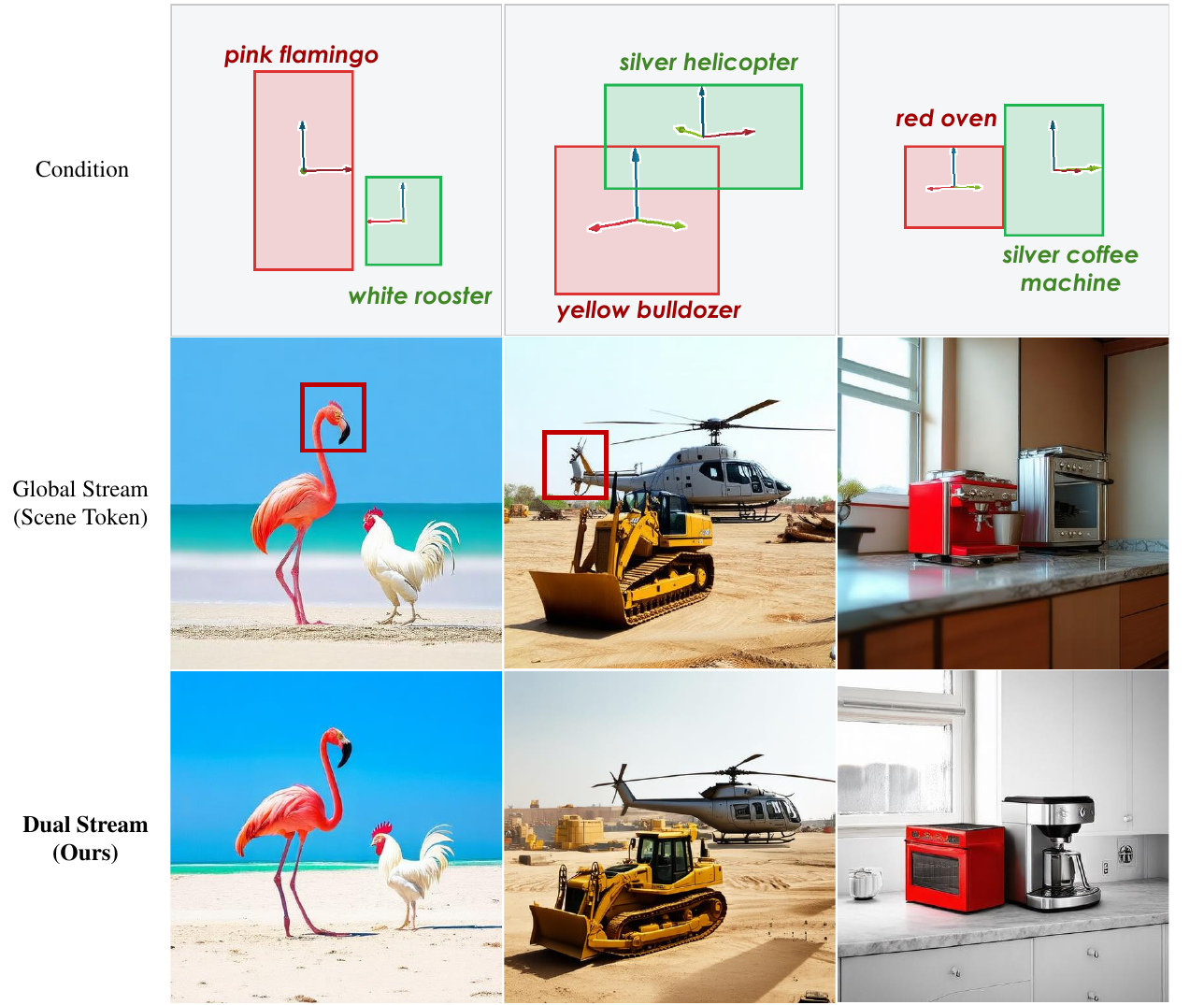}
  \caption{Qualitative ablation results of relying solely on the global stream.}
  \Description{A qualitative comparison matrix with three columns: Condition, Global Stream (Scene Token), and Dual Stream (Ours). It displays generated images for three prompts: a pink flamingo and a white rooster, a yellow bulldozer and a silver helicopter, and a red oven and a silver coffee machine. The Global Stream shows attribute leakage, while the Dual Stream cleanly separates the attributes.}
  \label{fig:ablation_qualitative_1}
\end{figure}

\begin{table}[t]
\centering
\caption{Quantitative ablation study of PoseAdapter.}
\label{tab:ablation}
\resizebox{\linewidth}{!}{
\begin{tabular}{ll ccc cc}
\toprule
\multirow{2}{*}{Method} & \multirow{2}{*}{Split} & \multicolumn{3}{c}{\textit{OrientLayout}} & \multicolumn{2}{c}{Overall} \\
\cmidrule(lr){3-5} \cmidrule(lr){6-7}
& & $Acc_{ls}$ (\%) $\uparrow$ & mIoU (\%) $\uparrow$ & $Acc_{BOV}$ (\%) $\uparrow$ & CLIP $\uparrow$ & O-CLIP $\uparrow$ \\
\midrule
\multirow{2}{*}{w/o Box Mask Loss} & Single & 97.56 & 87.32 & 90.55 & \multirow{2}{*}{0.342} & \multirow{2}{*}{0.309} \\
 & Multi & 86.84 & 79.06 &90.61  &  & \\
\addlinespace
\multirow{2}{*}{w/o $t$-Sampling Shift} & Single & 96.40 & 86.21 & 89.45& \multirow{2}{*}{0.341} & \multirow{2}{*}{0.307} \\
 & Multi & 87.86 & 78.44 & 89.91 & & \\
\addlinespace
\multirow{2}{*}{w/o Both Strategies} & Single & 94.79 & 82.66 & 86.81 & \multirow{2}{*}{0.323} & \multirow{2}{*}{0.300} \\
 & Multi & 86.63 & 77.60 & 87.13 & & \\
\midrule
Global Stream Only & Multi & 88.65 & 77.78 & 92.17 & 0.342 & 0.308 \\
Local Stream Only & Multi & 91.50 & 81.25 & 88.69 & 0.334 & 0.310 \\
\midrule
\multirow{2}{*}{\textbf{PoseAdapter (Ours)}} & Single & \textbf{98.42} & \textbf{88.30} & \textbf{92.97} & \multirow{2}{*}{\textbf{0.344}} & \multirow{2}{*}{\textbf{0.310}} \\
 & Multi & \textbf{91.67} & \textbf{81.72} & \textbf{93.22} & & \\
\bottomrule
\end{tabular}
}
\end{table}

\begin{figure}[t]
  \centering
  \includegraphics[width=0.98\linewidth]{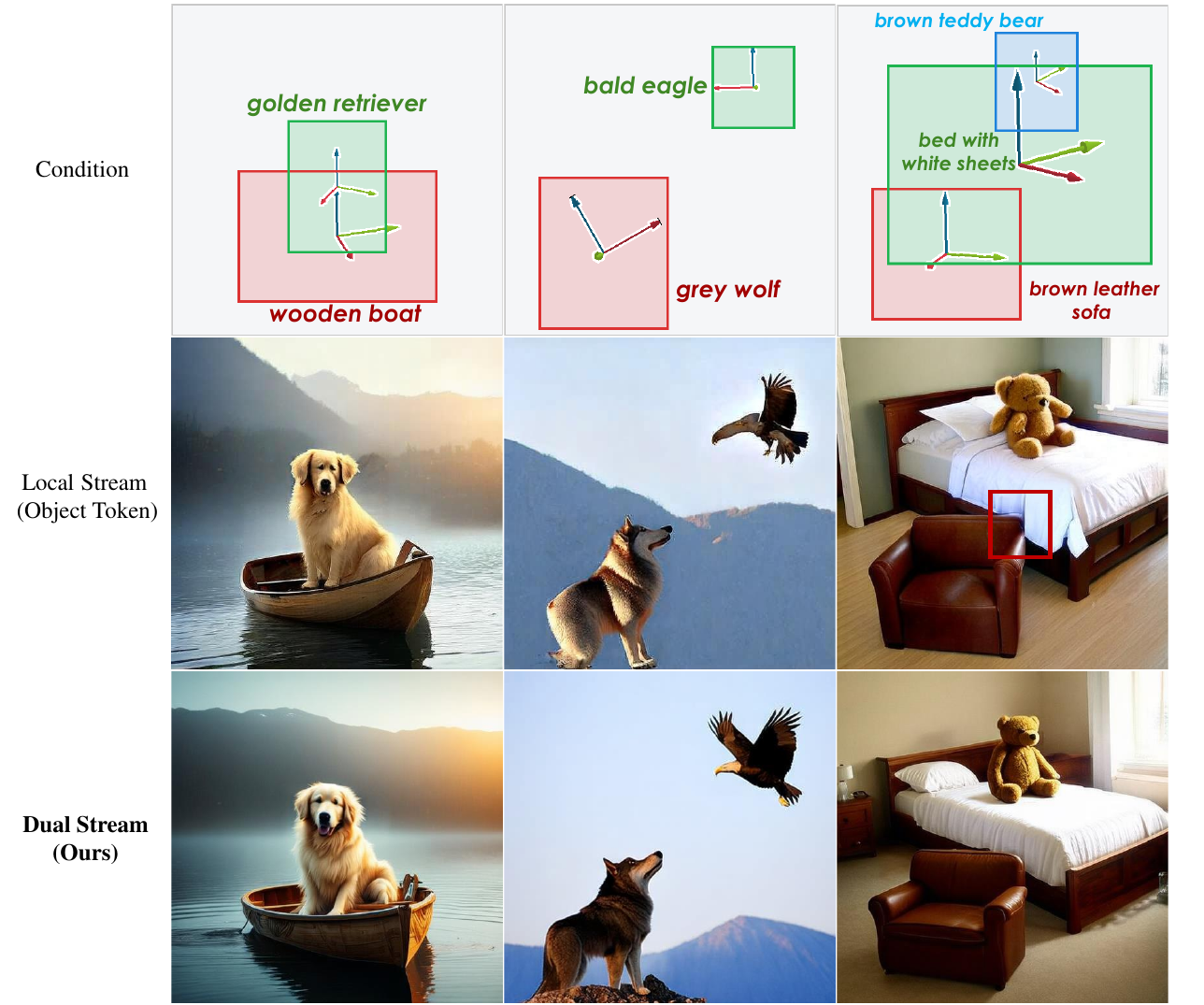}
  \Description{A qualitative comparison matrix with three columns: Condition, Local Stream (Object Token), and Dual Stream (Ours). It displays generated images for three prompts: a golden retriever on a wooden boat, a grey wolf and a bald eagle, and a bedroom scene with a sofa. The Local Stream shows disjointed composition and unnatural clipping (highlighted by a red box on the sofa), while the Dual Stream achieves photorealistic harmony.}
  \caption{Qualitative ablation results of relying solely on the local stream.}
  \label{fig:ablation_qualitative_2}
\end{figure}

\noindent\textbf{Effectiveness of the Context-Aware Dual-Stream.}
The global stream, with soft spatial constraints, captures inter-object relationships and consistent illumination essential for 3D orientation inference. However, it lacks explicit spatial routing and becomes vulnerable to attribute leakage where attention to one object bleeds into neighboring regions (Figure \ref{fig:ablation_qualitative_1}:
flamingo generates rooster's comb, helicopter absorbs bulldozer's hue). In contrast, the local stream employs strict box masking to achieve superior spatial accuracy and eliminate cross-object interference, yet by restricting the receptive field to
isolated regions, it deprives the model of global contextual information necessary for accurate 3D orientation inference, and therefore exhibits disjointed lighting and perspective misalignment (Figure \ref{fig:ablation_qualitative_2}: golden retriever and grey
wolf appear unnaturally pasted, brown leather sofa suffers clipping). This trade-off between spatial precision and orientation accuracy motivates our dual-stream design: by combining both streams, PoseAdapter enforces precise boundaries via the local
stream and achieves accurate orientation control by leveraging the global stream’s rich contexts.

\noindent\textbf{Effectiveness of Training Strategies.} We further ablate our proposed training objectives. Removing box mask loss or 
t-sampling shift degrades both spatial and orientational precision; removing both causes the largest drop (Table \ref{tab:ablation}). The two strategies synergize for robust grounding: the box loss up-weights foreground regions to prevent boundary violations, while the 
t-sampling shift prioritizes coarse geometric layout during high-noise Flow Matching steps.

\section{Conclusion}
\label{sec:conclusion}

We present PoseAdapter, a lightweight framework for 2.5D controllable generation that avoids dense 3D maps by anchoring objects with captions, 2D boxes, and 3D angles. A Context-Aware Dual-Stream Representation decouples masked local tokens from unmasked global tokens within the visual stream, eliminating attribute leakage while preserving inter-object coherence. Trained on our OrientLayout dataset, PoseAdapter surpasses prior arts in spatial, orientational, and visual fidelity. Currently limited to real-world priors, the model may struggle with extreme layouts; future work will augment synthetic data for enhanced robustness.

\section{Acnowledgments}
This work was supported by the Fundamental and Interdisciplinary Disciplines Breakthrough Plan of the Ministry of Education of China (No. JYB2025XDXM101) and 
the National Natural Science Foundation of China (project No. 62595773).

\bibliographystyle{ACM-Reference-Format}
\balance
\bibliography{references}

\end{document}